\documentclass[10pt,twocolumn,letterpaper]{article}

\usepackage{iccv}
\usepackage{times}
\usepackage{epsfig}
\usepackage{graphicx}
\usepackage{amsmath}
\usepackage{amssymb}
\usepackage{color}
\usepackage{xcolor}
\usepackage{enumerate}
\usepackage{enumitem}
\usepackage{pifont}
\newcommand{\cmark}{\text{\ding{51}}}
\newcommand{\xmark}{\text{\ding{55}}}
\definecolor{mypink}{RGB}{219, 48, 122}

\usepackage[colorlinks]{hyperref}

\iccvfinalcopy 

\def\iccvPaperID{3877} 

\ificcvfinal\fi

\begin{document}
\setlength{\abovedisplayskip}{4pt}
\setlength{\belowdisplayskip}{4pt}
\title{3D Human Pose Estimation with Spatial and Temporal Transformers}

\author{Ce Zheng$^{1}$, Sijie Zhu$^{1}$, Matias Mendieta$^{1}$, Taojiannan Yang$^{1}$, Chen Chen$^{1}$, Zhengming Ding$^{2}$\\
$^1$Center for Research in Computer Vision, University of Central Florida, USA\\
$^2$Department of Computer Science, Tulane University, USA\\
{\tt\small \{cezheng,sizhu,mendieta,taoyang1122\}@knights.ucf.edu;}\\
{\tt\small chen.chen@crcv.ucf.edu;zding1@tulane.edu}
}

\maketitle
\ificcvfinal\thispagestyle{empty}\fi

\begin{abstract}
   Transformer architectures have become the model of choice in natural language processing and are now being introduced into computer vision tasks such as image classification, object detection, and semantic segmentation. However, in the field of human pose estimation, convolutional architectures still remain dominant. In this work, we present \textbf{PoseFormer}, a purely transformer-based approach for 3D human pose estimation in videos without convolutional architectures involved. 
   Inspired by recent developments in vision transformers, we design a spatial-temporal transformer structure to comprehensively model the human joint relations within each frame as well as the temporal correlations across frames, then output an accurate 3D human pose of the center frame. We quantitatively and qualitatively evaluate our method on two popular and standard benchmark datasets: Human3.6M and MPI-INF-3DHP. Extensive experiments show that PoseFormer achieves state-of-the-art performance on both datasets. \textbf{Code is available at} \url{https://github.com/zczcwh/PoseFormer}
\end{abstract}

\section{Introduction}

Human pose estimation (HPE) aims to localize joints and build a body representation (\eg skeleton position) from input data such as images and videos. HPE provides geometric and motion information of the human body and can be applied to a wide range of applications (\eg human-computer interaction, motion analysis, healthcare). Current works generally can be divided into two classes: (1) direct estimation approaches, and (2) 2D-to-3D lifting approaches. Direct estimation methods~\cite{pavlakos2018ordinal, Moon_I2L_MeshNet} infer a 3D human pose from 2D images or video frames without intermediately estimating the 2D pose representation. 2D-to-3D lifting approaches \cite{Liu_2020_CVPR, chen2020anatomy, zeng2020srnet_ECCV,wang2020motion} infer 3D human pose from an intermediately estimated 2D pose. Benefiting from the excellent performance of state-of-the-art 2D pose detectors, 2D-to-3D lifting approaches generally outperform direct estimation methods. However, the mapping of these 2D poses to 3D is non-trivial; various potential 3D poses could be generated from the same 2D pose due to depth ambiguity and occlusion. To alleviate some of these issues and preserve natural coherence, many recent works have integrated temporal information from videos into their approaches. For example, \cite{ Liu_2020_CVPR, chen2020anatomy} utilize temporal convolutional neural networks (CNNs) to capture global dependencies from adjacent frames, and \cite{Hossain_2018_ECCV} uses recurrent architectures to similar effect. However, the temporal correlation window is limited for both of these architectures. CNN-based approaches typically rely on dilation techniques, which inherently have limited temporal connectivity, and recurrent networks are mainly constrained to simply sequential correlation.

Recently, the transformer \cite{Attention_is_All_You_Need} has become the \textit{de facto} model for natural language processing (NLP) due to its efficiency, scalability and strong modeling capabilities. Thanks to the self-attention mechanism of the transformer, global correlations across long input sequences can be distinctly captured. This makes it a particularly fitting architecture for sequence data problems, and therefore naturally extendable to 3D HPE. With its comprehensive connectivity and expression, the transformer provides an opportunity to learn stronger temporal representations across frames. 
However, recent works \cite{Dosovitskiy2020ViT, touvron2020deit} show that transformers require specific designs to achieve comparable performance with CNN counterparts for vision tasks. Specifically, they often require either extremely large scale training datasets \cite{Dosovitskiy2020ViT}, or enhanced data augmentation and regularization \cite{touvron2020deit} if applied to smaller datasets. Moreover, existing vision transformers have been limited primarily to image classification~\cite{Dosovitskiy2020ViT, touvron2020deit}, object detection~\cite{carion2020DETR, zhu2020deformableDETR}, and segmentation~\cite{ye2019segmentation, zheng2020rethinking}, but how to harness the power of transformers for 3D HPE remains unclear.

\begin{figure}[t]
\begin{center}
 \includegraphics[width=.85\linewidth]{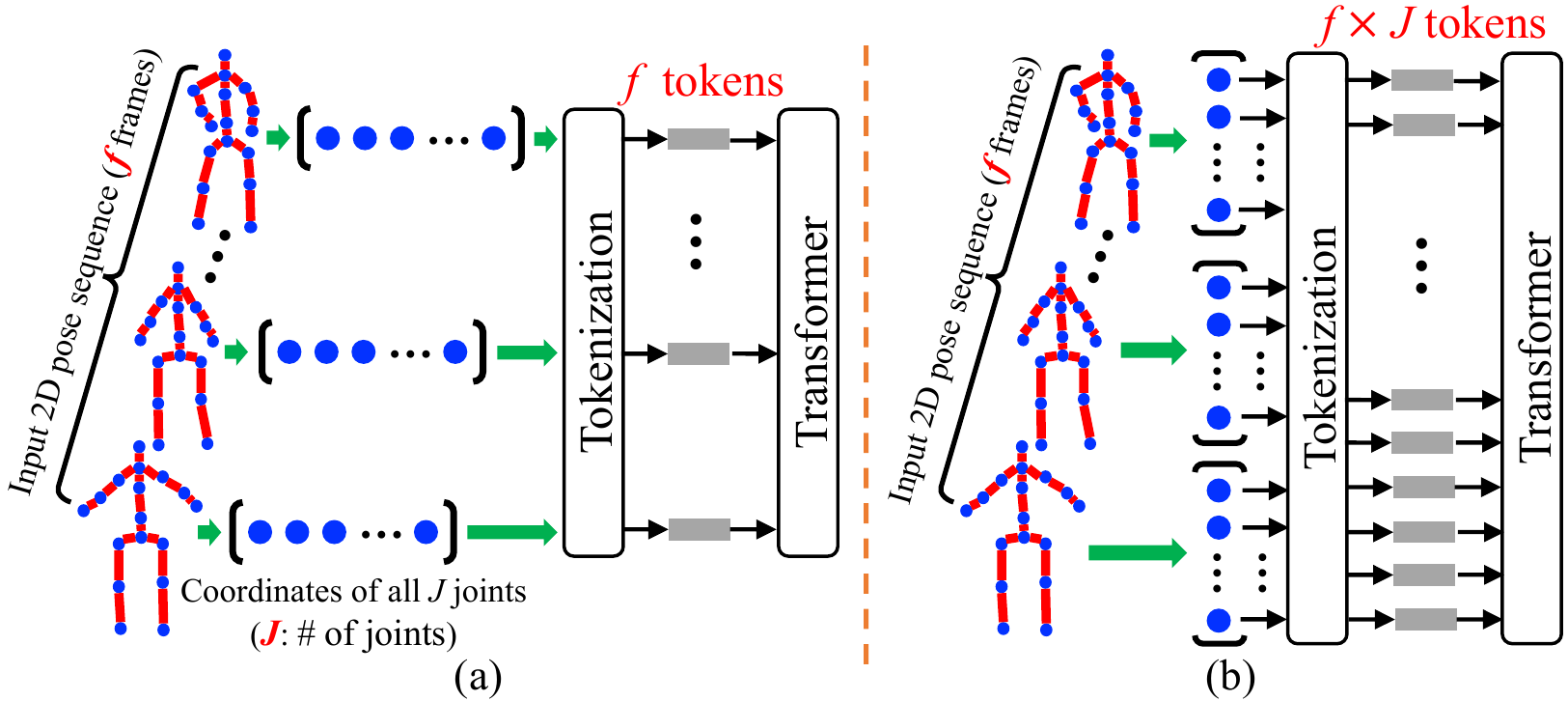}
\end{center}
\vspace{-10pt}
   \caption{Two baseline approaches.}
\label{fig:baseline}
\vspace{-15pt}
\end{figure}
To begin answering this question, we first directly apply the transformer on 2D-to-3D lifting HPE. In this case, we view the entire 2D pose for each frame in a given sequence as a token (Fig. \ref{fig:baseline}(a)). While this baseline approach is functional to an extent, it ignores the natural distinction of spatial relations (joint-to-joint),
leaving potential improvements on the table. A natural extension to this baseline is to instead view each 2D joint coordinate as a token, and provide an input formed with these joints from across all frames of the sequence (Fig. \ref{fig:baseline}(b)). However, in this case, the number of tokens becomes increasingly large when long frame sequences are used (up to 243 frames and 17 joints per frame is common in 3D HPE, the number of tokens would be 243$\times$17$=$4131). Since the transformer computes direct attention with each token to another, the memory requirement of the model approaches an unreasonable level.

Therefore, as an effective solution to these challenges,
we propose \textbf{PoseFormer}, the first pure transformer network for 2D-to-3D lifting HPE in videos. PoseFormer directly models the spatial and temporal aspects with distinct transformer modules for both dimensions. 
Not only does PoseFormer produce strong representations across the spatial and temporal elements, it does so without inducing enormous token counts for long input sequences. 
On a high level, PoseFormer simply takes a sequence of detected 2D poses from an off-the-shelf 2D pose estimator, and outputs the 3D pose for the center frame. More specifically, we build a spatial transformer module to encode local relationships between the 2D joints in each frame. The spatial self-attention layers consider the position information of 2D joints and return a latent feature representation for that frame. Next, our temporal transformer module analyzes global dependencies between each spatial feature representation, and generates an accurate 3D pose estimation. 

Experimental evaluations on two popular 3D HPE benchmarks, Human3.6M \cite{Human3.6M} and MPI-INF-3DHP \cite{MPIINF}, show that PoseFormer achieves state-of-the-art performance on both datasets. We visualize our estimated 3D pose compared with the state-of-the-art approach, and find that PoseFormer produces smoother and more reliable results. Also, visualizations and analyses of PoseFormer's attention maps are provided in the ablation study to understand the internal workings of our model and demonstrate its effectiveness.
Our \textbf{contributions} are three-fold:
\setlist{nolistsep}
\begin{itemize}[noitemsep,leftmargin=*]  
\item We propose the first pure transformer-based model, PoseFormer, for 3D HPE under the category of 2D-to-3D lifting.  

\item We design an effective Spatial-Temporal Transformer model, where the spatial transformer module encodes local relationships between human body joints, and the temporal transformer module captures the global dependencies across frames in the entire sequence.

\item Without bells and whistles, our PoseFormer model achieves state-of-the-art results on both Human3.6M and MPI-INF-3DHP datasets. 
   
\end{itemize}

\begin{figure*}[htp]
\vspace{-5pt}
  \centering
  \includegraphics[width=0.90\linewidth]{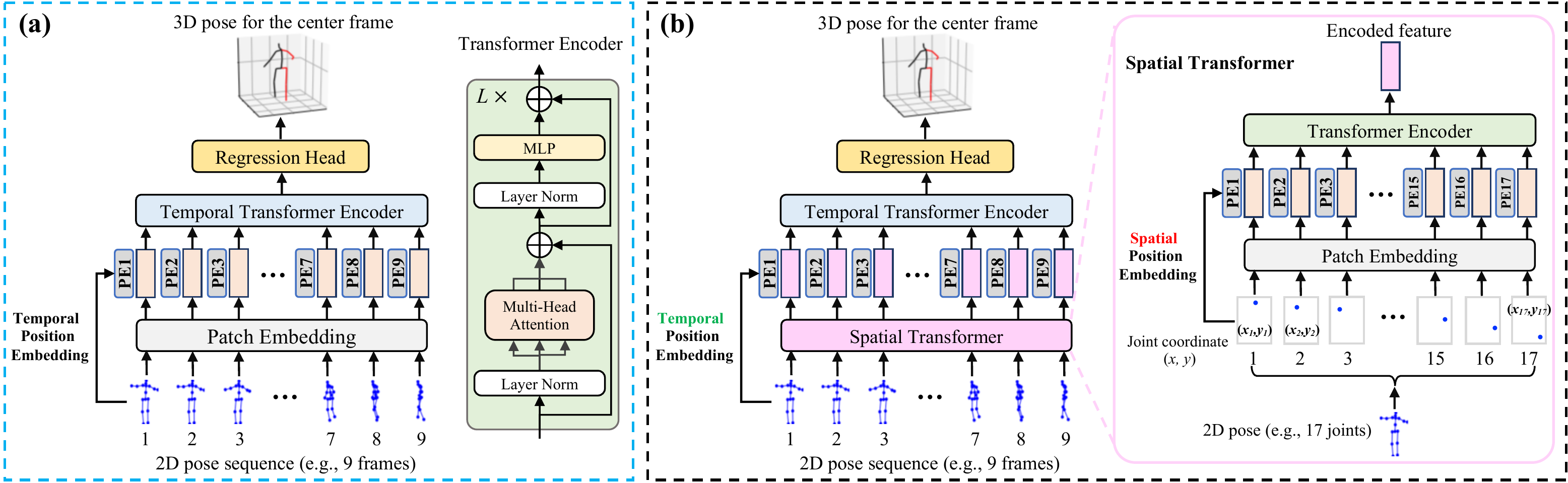}
  \caption{(a) Temporal transformer baseline. (b) Spatial-temporal transformer (PoseFormer) architecture, which consists of three modules. A spatial transformer module for extracting features with considering joints correlations of each individual skeleton. A temporal transformer module for learning global dependencies of entire sequence. A regression head module regresses the final 3D pose of the center frame. The illustration of the transformer encoder is followed by ViT \cite{Dosovitskiy2020ViT}.  }
  \label{fig:architecture}
  \vspace{-10pt}
\end{figure*}
\section{Related Works}


Here we specifically summarize 3D single-person-single-view HPE methods. 
Direct estimation approaches infer 3D human pose from 2D images without intermediately estimating 2D pose representation. 2D-to-3D lifting approaches utilize the 2D pose as input to generate the corresponding 3D pose, which is more popular among state-of-the-art methods in this domain. Any off-the-shelf 2D pose estimator can be effectively compatible with these methods.
Our proposed method, PoseFormer, also follows the 2D-to-3D lifting pipeline, and therefore we will focus mainly on such methods in this section.


\textbf{2D-to-3D Lifting HPE.}
2D-to-3D lifting approaches leverage 2D poses estimated from input images or video frames. OpenPose \cite{cao2017realtime}, CPN \cite{chen2018cascaded}, AlphaPose \cite{fang2017rmpe}, and HRNet \cite{sun2019deep} have been extensively used as the 2D pose detectors. Based on this intermediate representation, the 3D pose can be generated with a variety of methods. Martinez \etal \cite{martinez_2017_3dbaseline} proposed a simple and effective fully connected residual network to regress 3D joint locations based on the 2D joint locations from just a single frame.
However, instead of estimating 3D human pose from monocular images, videos can provide temporal information to improve accuracy and robustness \cite{Zhou_2017_ICCV, Dabral_2018_ECCV, pavllo2019, Cheng2019occlusionaware, Cai2019Spatial_Temporal, Zhang_2020_CVPR_Object_Occluded, wang2020motion}. Hossain and Little \cite{Hossain_2018_ECCV} proposed a recurrent neural network using Long Short-Term Memory (LSTM) cells to exploit temporal information in the input sequence. Several works \cite{Dabral_2018_ECCV, Cai2019Spatial_Temporal,Li2019boosting} utilized spatial-temporal relationships and constraints such as bone-length and left-right symmetry to improve performance. Pavllo \etal \cite{pavllo2019} introduced a temporal convolution network to estimate 3D pose over 2D keypoints from consecutive 2D sequences. Based on \cite{pavllo2019}, Chen \etal \cite{chen2020anatomy} added a bone direction module and bone length module to ensure temporal consistency across video frames, and Liu \etal \cite{Liu_2020_CVPR} utilized an attention mechanism to recognize significant frames. 
However, the previous state-of-the-art methods (\eg \cite{Liu_2020_CVPR, chen2020anatomy}) rely on dilated temporal convolutions to capture global dependencies, which are inherently limited in temporal connectivity. Additionally, the majority of these works \cite{Liu_2020_CVPR, chen2020anatomy, Hossain_2018_ECCV, pavllo2019} project the joint coordinates to a latent space using simple operations, without considering the kinematic correlations of human joints.

\textbf{GNNs in 3D HPE}. Naturally, a human pose can be represented as a graph where the joints are the nodes and the bones are the edges. Graph Neural Networks (GNNs) have also been applied to the 2D-to-3D pose lifting problem and provided promising performance \cite{Ci2019, Zhao_2019_Semantic_Graph, Liu_2020_ECCV_weight_sharing}. 
Ci \etal \cite{Ci2019} proposed a framework, named Locally Connected Networks (LCNs), which leverages both fully connected networks and GNN operations to encode the relationship between local joint neighborhoods. Zhao \etal \cite{Zhao_2019_Semantic_Graph} tackled a limitation of Graph Convolutional Network \cite{gcn} (GCN) operations, specifically how the weight matrix is shared across nodes. The semantic graph convolution operation was introduced to learn channel-wise weights for edges.

For our PoseFormer, the transformer can be viewed as a type of graph neural network with a unique, and often advantageous, graph operation. Specifically, a transformer encoder module essentially forms a fully-connected graph, where the edge weights are computed using input-conditioned, multi-headed self-attention. The operation also includes the normalization of node features, a feed-forward aggregator across attention head outputs, and residual connections which enable it to scale effectively with stacked layers. Such an operation can be advantageous in comparison to other graph operations. For example, the strength of the connection between nodes is determined by the self-attention mechanism of the transformer, rather than predefined through an adjacency matrix as with the typical GCN-based formulations employed in this task. This allows the model flexibility to adapt the relative importance of joints to each other with each input pose. Additionally, the comprehensive scaling and normalization components of the transformer are likely advantageous in mitigating the over-smoothing effect that troubles many GNN operation variants when numerous layers are stacked together \cite{gnnNorm}.

\textbf{Vision Transformers.} 
Recently, there is an emerging interest in applying transformers to vision tasks \cite{khan2021transformers, han2020survey}. Carion \etal \cite{carion2020DETR} presented a DEtection TRansformer (DETR) for object detection and panoptic segmentation. Dosovitskiy \etal \cite{Dosovitskiy2020ViT} proposed a pure transformer architecture, Vision Transformer (ViT), which achieves state-of-the-art performance on image classification. However, ViT was trained on large-scale datasets ImageNet-21k and JFT-300M that requires huge computation resources. Then, a data-efficient image transformer (DeiT) \cite{touvron2020deit} was proposed which builds upon the ViT with knowledge distillation. For regression problems such as HPE, Yang \etal \cite{yang2020transpose} proposed a transformer network, Transpose, which only estimates 2D pose from images. Lin \etal \cite{lin2020METRO} combined CNNs with transformer networks in their method METRO (MEsh TRansfOrmer) to reconstruct the 3D pose and mesh vertices from a single image. In contrast to our approach, METRO falls under the category of direct estimation. Also, temporal consistency is neglected in METRO, which limits the robustness of its estimations. Our spatial-temporal transformer architecture exploits keypoint correlation in each frame and preserves natural temporal coherence in videos.

 
\section{Method}

We follow the same 2D-to-3D lifting pipeline for 3D HPE in videos as \cite{martinez_2017_3dbaseline, pavllo2019, Liu_2020_CVPR, chen2020anatomy}. The 2D pose of each frame is obtained by an off-the-shelf 2D pose detector, then 2D pose sequences of consecutive frames are used as input for estimating the 3D pose of the center frame. Compared to the previous state-of-the-art models, which are based on CNNs, we produce a highly competitive \textbf{convolution-free} transformer network. 


\subsection{Temporal Transformer Baseline} \label{baseline}
As a baseline application of a transformer in 2D-to-3D lifting, we treat each 2D pose as an input token and employ a transformer to capture global dependencies among the inputs as illustrated in Fig. \ref{fig:architecture}(a). We will refer to each input token as a patch, similar in terminology to ViT \cite{Dosovitskiy2020ViT}. For the input sequence $X\in \mathbb{R} ^{f \times (J \cdot 2)}$, $f$ is the number of frames of the input sequence, $J$ is the number of joints of each 2D pose, and 2 indicates joint's coordinate in 2D space. \{$\mathbf{x}^i  \in \mathbb{R} ^{1 \times (J \cdot 2) } | i = 1,2,\dots f$\} indicates the input vector of each frame. The patch embedding is a trainable linear projection layer to embed each patch to a high dimensional feature. 
The transformer network utilizes positional embeddings to retain positional information of the sequence. The procedure can be formulated as:
\begin{align}
\small
    Z_0 = [\mathbf{x}^1 E; \ \mathbf{x}^2 E ; \ \dots; \ \mathbf{x}^f E ] + E_{pos}.
\end{align}

After embedding through a linear projection matrix $E \in \mathbb{R} ^{(J\cdot 2) \times C}$ and summing with the positional embedding $E_{pos}\in \mathbb{R} ^{f \times C}$, the input sequence $X\in \mathbb{R} ^{f \times (J \cdot 2)}$ becomes $Z_0\in \mathbb{R} ^{f \times C}$, where $C$ is the embedding dimension. 
$Z_0$ is sent to the Temporal Transformer Encoder.

As the core function of the transformer, self-attention is designed to relate different positions of the input sequence with embedded features. Our transformer encoder is composed of Multi-head Self Attention blocks with multilayer perceptron (MLP) blocks as in \cite{Dosovitskiy2020ViT}. LayerNorm is applied before every block and residual connections are applied after every block \cite{wang2019layernorm, baevski2018adaptive}.



\textbf{Scaled Dot-Product Attention} can be described as a mapping function that maps a query matrix $Q$, key matrix $K$ and value matrix $V$ to an output attention matrix. $Q, K, V\in \mathbb{R} ^{N \times d}$, where $N$ is the number of vectors in the sequence and $d$ is the dimension. A scaling factor of $\frac{1}{\sqrt{d}}$ is utilized within this attention operation for appropriate normalization, preventing extremely small gradients when large values of $d$ lead dot products to grow large in magnitude. Thus the output of the scaled dot-product attention can be expressed as:
\begin{align}
\small
    {\rm Attention}(Q,K,V) = {\rm Softmax}(QK^\top/ \sqrt{d})V.
\end{align}

In our temporal transformer, $d = C$ and $N = f$. The $Q$, $K$ and $V$ are computed from the embedded feature ${Z}\in \mathbb{R} ^{f \times C}$ by linear transformations $W_Q$, $W_K$ and $W_V$ $\in \mathbb{R} ^{C \times C}$:
\begin{align}
\small
    Q = {Z}W_Q, \quad K = {Z}W_K, \quad V = {Z}W_V.
\end{align}

\textbf{Multi-head Self Attention Layer} (MSA) utilizes multiple heads to model the information jointly from various representation subspaces with different positions. Each head applies scaled dot-product attention in parallel. The MSA output will be the concatenation of $h$ attention head outputs.
\begin{align}
\small
    {{\rm MSA} (Q,K,V)} 
    &= {\rm Concat} (H_1, H_2, \dots, H_h) W_{out} \\  {\rm where} \quad H_i &= {\rm Attention} (Q_i,K_i,V_i), i \in [1,...,h]
\end{align}

The Temporal Transformer Encoder structure of $L$ layers given our embedded feature $Z_0\in \mathbb{R} ^{f \times C}$ can be represented as follows: 
\begin{align}
\small
    Z_{\ell}^{'} &= {\rm MSA}({\rm LN}(Z_{{\ell}-1}))+ Z_{{\ell}-1},  & \ell&= 1,2 \dots L \\
    Z_{\ell} &= {\rm MLP}({\rm LN}(Z_{\ell}^{'})) + Z_{\ell}^{'}, & \ell&= 1,2, \dots L \\
    Y &= {\rm LN}(Z_{L}),
\end{align}
where $\rm LN(\cdot)$ denotes the layer normalization operator (same as in ViT). The temporal transformer encoder consists of $L$ identical layers and the encoder output $Y\in \mathbb{R} ^{f \times C}$ keeps the same size as encoder input $Z_0 \in \mathbb{R} ^{f \times C}$.




In order to predict the 3D pose of center frame, the encoder output $Y\in \mathbb{R} ^{f \times C}$ is shrunk to a vector $\mathbf{y} \in \mathbb{R} ^{1 \times C}$ by taking the average in the frame dimension. Finally, an MLP block will regress the output to $\mathbf{\mathbf{y}} \in \mathbb{R} ^{1 \times (J \cdot 3)}$, which is the 3D pose of the center frame. 

\subsection{PoseFormer: Spatial-Temporal Transformer}
We observe that the temporal transformer baseline mainly focuses on global dependencies between frames in the input sequence. 
The patch embedding, a linear transformation, is utilized to project joint coordinates to a hidden dimension. However, the kinematic information between local joint coordinates is not strongly represented in the temporal transformer baseline because \textit{a simple linear projection layer} is not able to learn attention information. One potential workaround is to view each joint coordinate as an individual patch and feeding the joints from all frames as input to the transformer (see Fig. \ref{fig:baseline}(b)). However, the number of patches would increase rapidly (frames $f$ multiplied by the number of joint $J$), resulting in a model computational complexity of $O((f \cdot J)^2)$. For example, if we use 81 frames and 17 joints for each 2D pose, the number of patches would be 1377 (ViT model uses 576 patches (input size = $384 \times 384$, patch size = $16 \times 16$)). 

To learn local joint correlations efficiently, we employ two separated transformers for spatial and temporal information, respectively. As shown in Fig. \ref{fig:architecture}(b), PoseFormer consists of three modules: spatial transformer module, temporal transformer module, and regression head module.

\textbf{Spatial Transformer Module.} The spatial transformer module is to extract a high dimensional feature embedding from a single frame. Given a 2D pose with $J$ joints, we consider each joint (\ie two coordinates) as a patch and follow the general vision transformer pipeline to perform the feature extraction among all patches. First, we map the coordinate of each joint to a high dimension with a trainable linear projection, which is referred to as the spatial patch embedding. We sum this with the learnable spatial positional embedding \cite{Dosovitskiy2020ViT} $E_{SPos} \in \mathbb{R} ^{J \times c}$, and therefore the input $\mathbf{x}_i \in \mathbb{R} ^{1 \times (J \cdot 2)}$ of the $i$-{th} frame becomes $z_{0}^i \in \mathbb{R} ^{J \times c}$, where 
2 indicates 2D coordinate in each frame and $c$ is the spatial embedding dimension. 
The resulting joint sequence of features  $z_{0}^i$ are fed into the spatial transformer encoder which applies the self-attention mechanism to integrate information across all joints. For the $i$-{th} frame, the output of spatial transformer encoder with $L$ layers will be $z_{L}^i \in \mathbb{R} ^{J \times c} $.

\textbf{Temporal Transformer Module.}
Since the spatial transformer module encodes high dimensional features for each individual frame, the goal for the temporal transformer module is to model dependencies across the sequence of frames. For the $i$-{th} frame, the output of the spatial transformer 
$z_{L}^i \in \mathbb{R} ^{J \times c}$ is flattened as a vector $\mathbf{z}^i  \in \mathbb{R} ^{1 \times (J \cdot c)}$.
We then concatenate these vectors \{$\mathbf{z}^1, \mathbf{z}^2, \dots, \mathbf{z}^f$\} from the $f$ input frames as $Z_{0} \in \mathbb{R} ^{f \times (J\cdot c)}$. 
Before the temporal transformer module, we add the learnable temporal positional embedding \cite{Dosovitskiy2020ViT} $E_{TPos} \in \mathbb{R} ^{f \times (J \cdot c)}$ to retain frame position information. For the temporal transformer encoder, we apply the same architecture as the spatial transformer encoder, which consists of multihead self-attention blocks and MLP blocks. The output of the temporal transformer module is  $Y \in \mathbb{R} ^{f \times (J \cdot c)}$. 

\textbf{Regression Head.}
Since we use a sequence of frames to predict the 3D pose of the center frame, the output of the temporal transformer module $Y \in \mathbb{R} ^{f \times (J\cdot c)}$ needs to be reduced to $\mathbf{y} \in \mathbb{R} ^{1 \times (J\cdot c)}$. We apply a weighted mean operation (with learned weights) on the frame dimension to achieve this. Finally, a simple MLP block with Layer norm and one linear layer returns output $\mathbf{y} \in \mathbb{R} ^{1 \times (J\cdot 3)}$ which is the predicted 3D pose of the center frame.

\textbf{Loss Function.} To train our spatial-temporal transformer model, we apply the standard MPJPE (Mean Per Joint Position Error) loss to minimize the error between the predicted and ground truth pose as
\begin{align}
\small
    \mathcal{L} = \frac{1}{J}\Sigma_{k=1}^{J}\| p_k - {\hat{p}}_k \|_2,
\end{align}
where $p_k$ and ${\hat{p}}_k$ are the ground truth and estimated 3D joint locations of the $k$-{th} joint, respectively.

\section{Experiments}

\subsection{Datasets and Evaluation Metrics}

We evaluate our model on two commonly used 3D HPE datasets, Human3.6M \cite{Human3.6M} and MPI-INF-3DHP \cite{MPIINF}.

\textbf{Human3.6M} \cite{Human3.6M} is the most widely used indoor dataset for 3D single person HPE. There are 11 professional actors performing 17 actions such as sitting, walking, and talking on the phone. Videos of each subject were recorded from 4 different views in an indoor environment. This dataset contains 3.6 million video frames with 3D ground truth annotation captured by an accurate marker-based motion capture system. Following previous works \cite{pavllo2019,Liu_2020_CVPR,chen2020anatomy}, we adopt the same experiment setting: all 15 actions are used for training and testing, the model is trained on five sections (S1, S5, S6, S7, S8) and tested on two subjects (S9 and S11).

\textbf{MPI-INF-3DHP} \cite{MPIINF} is a more challenging 3D pose dataset. It contains both constrained indoor scenes and complex outdoor scenes. There are 8 actors performing 8 actions from 14 camera views which cover a greater diversity of poses. MPI-INF-3DHP provides a test set of 6 subjects with different scenes. We follow the setting in \cite{lin2019trajectory,chen2020anatomy,wang2020motion}.

For the Human3.6M dataset, we use the most common evaluation metrics (MPJPE and P-MPJPE)~\cite{zheng2020deep} to evaluate the performance of our estimation with the ground truth 3D pose. MPJPE (Mean Per Joint Position Error) is computed as the mean Euclidean distance between the estimated joints and the ground truth in millimeters; we refer to MPJPE as Protocol 1. P-MPJPE is the MPJPE after rigid alignment by post-processing between the estimated 3D pose and the ground truth and it is more robust to individual joint prediction failure. We refer to P-MPJPE as Protocol 2. 

For the MPI-INF-3DHP dataset, we use MPJPE, Percentage of Correct Keypoint (PCK) within the 150$mm$ range \cite{lin2019trajectory,chen2020anatomy,wang2020motion}, and Area Under Curve (AUC).

\subsection{Implementation Details}
We implemented our proposed method with Pytorch \cite{PyTorch}. Two NVIDIA RTX 3090 GPUs were used for training and testing. We chose three different frame sequence lengths when conducting our experiments, \ie $f$ = 9, $f$ = 27, $f$ = 81. The details about number of frames with results are discussed in the ablation studies (Sec. \ref{Ablation}). We apply pose flipping horizontally as data augmentation both in training and testing following \cite{pavllo2019, Liu_2020_CVPR, chen2020anatomy}. We train our model using the Adam \cite{kingma2014adam} optimizer for 130 epochs with weight decay of 0.1. We adopt an exponential learning rate decay schedule with the initial learning rate of 2e-4 and decay factor of 0.98 of each epoch. We set the batch size to 1024 and employ stochastic depth \cite{huang2016stochastic} with a rate of 0.1 for transformer encoder layers. 
For the 2D pose detector, we use the cascaded pyramid network (CPN) \cite{chen2018cpn} on Human3.6M following \cite{pavllo2019,Liu_2020_CVPR,chen2020anatomy}, and we use the ground truth 2D pose as input for MPI-INF-3DHP following \cite{mehta2017vnect,lin2019trajectory}.

\begin{table*}[]
\scriptsize
\centering
  \caption{Quantitative comparison of Mean Per Joint Position Error between the estimated 3D pose and the ground truth 3D pose on Human3.6M under Protocols 1\&2 using the detected 2D pose as input. Top-table: results under Protocol 1 (MPJPE). Bottom-table: results under Protocol 2 (P-MPJPE). $f$ denotes the number of input frames used in each method, $*$ indicates that the input 2D pose is detected by the cascaded pyramid network (CPN), and $\dagger$ denotes a Transformer-based model. (\textcolor{red}{Red}: best; \textcolor{blue}{Blue}: second best)}
  \resizebox{\linewidth}{!}{
  \begin{tabular}{lccccccccccccccccc}

\multicolumn{1}{c|}{\textbf{Protocol 1}}       &  \multicolumn{1}{c|}{}      & Dir. & Disc. & Eat. & Greet & Phone & Photo & Pose & Purch. & Sit  & SitD. & Somke & Wait & WalkD. & Walk & \multicolumn{1}{c|}{WalkT.} & Average \\ \hline
\multicolumn{1}{l|}{Dabral \etal \cite{dabral2018learning}}   & \multicolumn{1}{c|}{ECCV'18}      & 44.8 & 50.4  & 44.7 & 49.0  & 52.9  & 61.4  & 43.5 & 45.5   & 63.1 & 87.3  & 51.7  & 48.5 & 52.2   & 37.6 & \multicolumn{1}{c|}{41.9}   & 52.1    \\
\multicolumn{1}{l|}{Cai \etal \cite{Cai2019Spatial_Temporal} ($f=7$)}    & \multicolumn{1}{c|}{ICCV'19 }    & 44.6 & 47.4  & 45.6 & 48.8  & 50.8  & 59.0  & 47.2 & 43.9   & 57.9 & 61.9  & 49.7  & 46.6 & 51.3   & 37.1 & \multicolumn{1}{c|}{39.4}   & 48.8    \\
\multicolumn{1}{l|}{Pavllo \etal \cite{pavllo2019} ($f=243$)*}  & \multicolumn{1}{c|}{CVPR'19 } & 45.2 & 46.7  & 43.3 & 45.6  & 48.1  & 55.1  & 44.6 & 44.3   & 57.3 & 65.8  & 47.1  & 44.0 & 49.0   & 32.8 & \multicolumn{1}{c|}{33.9}   & 46.8    \\
\multicolumn{1}{l|}{Lin \etal \cite{lin2019trajectory} ($f=50$)}   & \multicolumn{1}{c|}{BMVC'19}     & 42.5 & 44.8  & 42.6 & 44.2  & 48.5  & 57.1  & 52.6 & \color{blue}41.4   & 56.5 & 64.5  & 47.4  & 43.0 & 48.1   & 33.0 & \multicolumn{1}{c|}{35.1}   & 46.6    \\
\multicolumn{1}{l|}{Yeh \etal \cite{yeh2019chirality} }       & \multicolumn{1}{c|}{NIPS'19}       & 44.8 & 46.1  & 43.3 & 46.4  & 49.0  & 55.2  & 44.6 & 44.0   & 58.3 & 62.7  & 47.1  & 43.9 & 48.6   & 32.7 & \multicolumn{1}{c|}{33.3}   & 46.7    \\
\multicolumn{1}{l|}{Liu \etal \cite{Liu_2020_CVPR} ($f=243$)*}   & \multicolumn{1}{c|}{CVPR'20}  &   41.8 & 44.8  & 41.1 & 44.9  & 47.4  & 54.1  & 43.4 & 42.2   & 56.2 & 63.6  & \color{red}45.3  & 43.5 & \color{blue}45.3   & \color{red}31.3 & \multicolumn{1}{c|}{\color{blue}32.2}   & 45.1    \\
\multicolumn{1}{l|}{SRNet \cite{zeng2020srnet_ECCV} *}       & \multicolumn{1}{c|}{ECCV'20}    & 46.6 & 47.1  & 43.9 & \color{red}41.6  & \color{red}45.8  & \color{red}49.6  & 46.5 & \color{red}40.0   & \color{blue}53.4 & 61.1  & 46.1  & 42.6 & \color{red}43.1   & \color{blue}31.5 & \multicolumn{1}{c|}{32.6}   & 44.8    \\
\multicolumn{1}{l|}{UGCN \cite{wang2020motion} ($f=96$)}    & \multicolumn{1}{c|}{ECCV'20}  &  \color{red} 41.3 & \color{blue}43.9  & 44.0 & \color{blue}42.2  & 48.0  & 57.1  &  \color{blue}42.2 & 43.2   & 57.3 & 61.3  & 47.0  & 43.5 & 47.0  & 32.6 & \multicolumn{1}{c|}{\color{red}31.8}   & 45.6    \\
\multicolumn{1}{l|}{Chen \etal \cite{chen2020anatomy} ($f=81$)*}  & \multicolumn{1}{c|}{TCSVT'21}  & 42.1 & \color{blue}43.8  &  \color{blue}41.0 & 43.8  & \color{blue}46.1  & 53.5  & 42.4 & 43.1   & 53.9 & \color{red}60.5  & 45.7  & \color{red}42.1 & 46.2   & 32.2 & \multicolumn{1}{c|}{33.8}   & \color{blue}44.6    \\
\multicolumn{1}{l|}{METRO \cite{lin2020METRO} ($f=1$) $\dagger$}    & \multicolumn{1}{c|}{CVPR'21}    & -    & -     & -    & -     & -     & -     & -    & -      & -    & -     & -     & -    & -      & -    & \multicolumn{1}{c|}{-}      & 54.0    \\
 \hline
\multicolumn{1}{l|}{\textbf{Baseline} ($f=81$)*$\dagger$}         & \multicolumn{1}{c|}{}     & 43.8 & 47.9 & 43.8  & 45.5  &  49.7 & 55.7 &  44.3  & 45.8 & 57.7  & 66.3  & 47.4 & 45.4  & 48.6 & 32.5 & \multicolumn{1}{c|}{33.8}   &  47.2   \\ 
\multicolumn{1}{l|}{\textbf{PoseFormer} ($f=81$)*$\dagger$}         & \multicolumn{1}{c|}{}    & \color{blue} 41.5 & 44.8  & \color{red} 39.8 & 42.5  & 46.5  & \color{blue}51.6  & \color{red}42.1 & 42.0   & \color{red}53.3 & \color{blue}60.7  & \color{blue}45.5  & \color{blue}43.3 & 46.1   & 31.8 & \multicolumn{1}{c|}{\color{blue}32.2}   & \color{red}44.3    \\ 
\hline

\hline

\hline
\multicolumn{1}{c|}{\textbf{Protocol 2}}       & \multicolumn{1}{c|}{}        & Dir. & Disc. & Eat. & Greet & Phone & Photo & Pose & Purch. & Sit  & SitD. & Somke & Wait & WalkD. & Walk & \multicolumn{1}{c|}{WalkT.} & Average \\ \hline
\multicolumn{1}{l|}{Pavlakos \etal \cite{pavlakos2018ordinal}}    & \multicolumn{1}{c|}{ CVPR'18}     & 34.7 & 39.8  & 41.8 & 38.6  & 42.5  & 47.5  & 38.0 & 36.6   & 50.7 & 56.8  & 42.6  & 39.6 & 43.9   & 32.1 & \multicolumn{1}{c|}{36.5}   & 41.8    \\
\multicolumn{1}{l|}{Hossain \etal \cite{Hossain_2018_ECCV}}     & \multicolumn{1}{c|}{ ECCV'18}     & 35.7   & 39.3  & 44.6 & 43.0  & 47.2  & 54.0  & 38.3 & 37.5   & 51.6 & 61.3  & 46.5  & 41.4 & 47.3   & 34.2 & \multicolumn{1}{c|}{39.4}   & 44.1    \\
\multicolumn{1}{l|}{Cai \etal \cite{Cai2019Spatial_Temporal} ($f=7$)}   & \multicolumn{1}{c|}{ICCV'19}     & 35.7 & 37.8  & 36.9 & 40.7  & 39.6  & 45.2  & 37.4 & 34.5   & 46.9 & 50.1  & 40.5  & 36.1 & 41.0   & 29.6 & \multicolumn{1}{c|}{32.3}   & 39.0    \\

\multicolumn{1}{l|}{Lin \etal \cite{lin2019trajectory} ($f=50$)}    & \multicolumn{1}{c|}{BMVC'19}    & \color{blue}32.5 & 35.3  & 34.3 & 36.2  & 37.8  & 43.0  & 33.0 & \color{blue}32.2   & 45.7 & 51.8  & 38.4  & 32.8 & 37.5   & 25.8 & \multicolumn{1}{c|}{28.9}   & 36.8    \\
\multicolumn{1}{l|}{Pavllo \etal \cite{pavllo2019} ($f=243$)*}  & \multicolumn{1}{c|}{CVPR'19 } & 34.1 & 36.1  & 34.4 & 37.2  & 36.4  & 42.2  & 34.4 & 33.6   & 45.0 & 52.5  & 37.4  & 33.8 & 37.8   & 25.6 & \multicolumn{1}{c|}{27.3}   & 36.5    \\
\multicolumn{1}{l|}{Liu \etal \cite{Liu_2020_CVPR} ($f=243$)*}   & \multicolumn{1}{c|}{CVPR'20 }  & \color{red}32.3 & \color{blue} 35.2  & \color{blue}33.3 & 35.8  & \color{blue}35.9  & \color{blue}41.5  & 33.2 & 32.7   & 44.6 & 50.9  & \color{blue}37.0  & \color{red}32.4 & 37.0   & \color{blue} 25.2 & \multicolumn{1}{c|}{27.2}   & 35.6    \\  		

\multicolumn{1}{l|}{UGCN \cite{wang2020motion} ($f=96$)}   & \multicolumn{1}{c|}{ECCV'20}   & 32.9 & \color{blue} 35.2  & 35.6 & \color{red}34.4  & 36.4  & 42.7  & \color{red}31.2 & 32.5   & 45.6 & 50.2  & 37.3  & 32.8 & \color{blue}36.3   & 26.0 & \multicolumn{1}{c|}{\color{red}23.9}   & \color{blue}35.5    \\
\multicolumn{1}{l|}{Chen \etal \cite{chen2020anatomy} ($f=81$)*}  & \multicolumn{1}{c|}{TCSVT'21 } & 33.1 & 35.3  & 33.4 & 35.9  & 36.1  & 41.7  & 32.8 & 33.3   & \color{red}42.6 & \color{blue}49.4  & \color{blue}37.0  & \color{blue}32.7 & 36.5   & 25.5 & \multicolumn{1}{c|}{27.9}   & 35.6    \\ \hline
\multicolumn{1}{l|}{\textbf{Baseline} ($f=81$)*$\dagger$}         & \multicolumn{1}{c|}{}     & 33.6 & 37.1 & 35.4  & 36.7  & 37.8  & 42.2 &  33.9  & 34.7 & 47.0  & 53.4  & 38.2 & 34.3  & 37.6 & 25.3 & \multicolumn{1}{c|}{27.8}   &  37.0   \\ 
\multicolumn{1}{l|}{\textbf{PoseFormer} ($f=81$)*$\dagger$}        & \multicolumn{1}{c|}{}    & \color{blue}32.5 & \color{red}34.8  & \color{red}32.6 & \color{blue}34.6  & \color{red}35.3  & \color{red}39.5  & \color{blue}32.1 & \color{red}32.0   & \color{blue}42.8 & \color{red}48.5  & \color{red}34.8  & \color{red}32.4 & \color{red}35.3   & \color{red}24.5 & \multicolumn{1}{c|}{\color{blue}26.0}   & \color{red}34.6    \\

\end{tabular}
}
\vspace{-10pt}
\label{tab: h36mp1}
\end{table*}

\subsection{Comparison with State-of-the-Art}
\textbf{Human3.6M}. We report all 15 action results of the test set (S9 and S11) in Table \ref{tab: h36mp1}. The last column provides the average performance on all of the test set. Following the 2D-to-3D lifting approach, we use the CPN network as the 2D pose detector, then use the detected 2D pose as input for training and testing. PoseFormer outperforms our baseline (\ie temporal transformer baseline in Sec. \ref{baseline}) by
a large margin (6.1\% and 6.4\%) under protocol 1 and protocol 2, respectively. This clearly demonstrates the advantage of using spatial transformer to expressively model the correlations between joints in each frame. PoseFormer yields the lowest average MPJPE of 44.3$mm$ under protocol 1 as shown in Table \ref{tab: h36mp1} (top). Comparing with the transformer-based method METRO \cite{lin2020METRO}, which ignores the temporal consistency since the 3D pose is estimated by a single image, PoseFormer reduces the MPJPE by approximately 18\%. For Protocol 2, we also obtain the best overall result as shown in Table \ref{tab: h36mp1} (bottom). Moreover, PoseFormer achieves more accurate pose predictions on difficult actions such as \textit{Photo}, \textit{SittingDown}, \textit{WalkDog}, and \textit{Smoke}. Unlike other simple actions, poses in these actions change more quickly and some long-distance frames have strong correlations. In this case, global dependencies play an important role, and the attention mechanisms of the transformer are particularly advantageous.

To further investigate the lower bound of our method, we directly use the ground truth 2D pose as input to alleviate error caused by noisy 2D pose data. The results are shown in Table \ref{tab: h36mgt}. The MPJPE is reduced from 44.3$mm$ to 31.3$mm$, about 29.7\% by using the clean 2D pose data. PoseFormer achieves the best score in 9 actions and the second best score in 6 actions. The average score is improved by approximately 2\% compared with SRNet \cite{zeng2020srnet_ECCV}.

\begin{figure}[htp]
  \centering
  \includegraphics[width=0.75\linewidth]{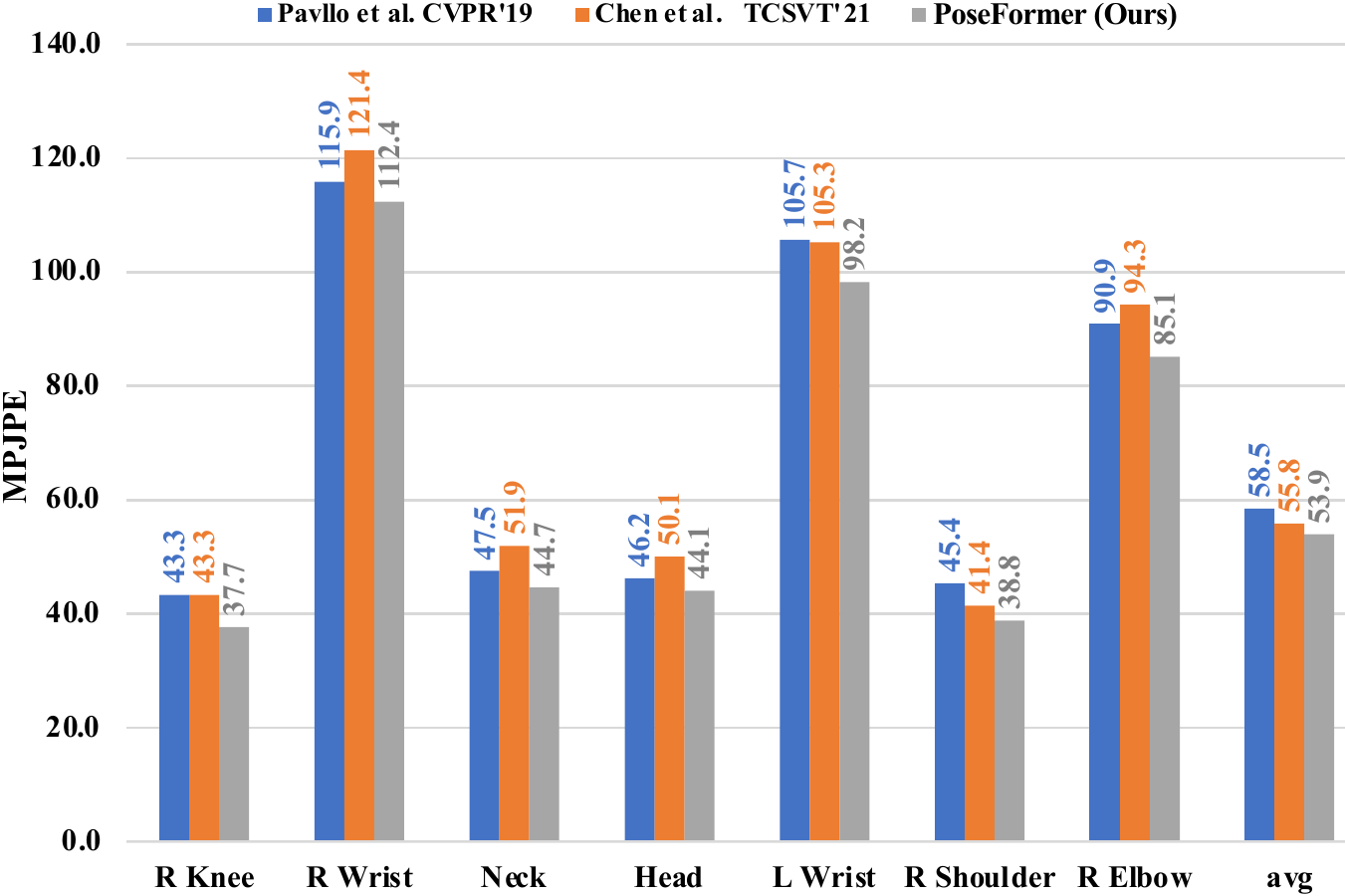}
  \caption{The average joint error comparison across all the frames of the Human3.6M test set S11 with the \textit{Photo} action. }
  \label{fig:eachmpjpe}
\end{figure}

In Fig. \ref{fig:eachmpjpe}, we also compare the MPJPE for some of the individual joints which have the largest errors on Human3.6M test set S11 with action \textit{Photo}. PoseFormer achieves better performance on these difficult joints than \cite{pavllo2019,chen2020anatomy}.   

\textbf{MPI-INF-3DHP}. Table \ref{tab:mpi} reports the quantitative results of PoseFormer with other methods on MPI-INF-3DHP. This dataset contains fewer training samples compared to Human3.6M, and some of the samples are taken from outdoor scenes. We use 2D poses of 9 frames as our model input due to the typically shorter sequence lengths of this dataset. Our method again achieves the best performances on all three evaluation metrics (PCK, AUC and MPJPE).

\begin{table*}[]
\scriptsize
\centering
  \caption{Quantitative comparison of Mean Per Joint Position Error between the estimated 3D pose and the ground truth 3D pose on Human3.6M dataset under Protocol 1 (MPJPE) using the \textbf{ground truth} 2D pose as input.  (\textcolor{red}{Red}: best; \textcolor{blue}{Blue}: second best)}
  \resizebox{\linewidth}{!}{
  \begin{tabular}{l|c|ccccccccccccccc|c}

\multicolumn{1}{c|}{GT   Protocol 1} & {} & Dir. & Disc. & Eat. & Greet & Phone & Photo & Pose & Purch. & Sit  & SitD. & Somke & Wait & WalkD. & Walk & WalkT. & Average \\ \hline
Hossain \etal \cite{Hossain_2018_ECCV}  & ECCV'18                     &
35.2 &	40.8 &	37.2 &	37.4 &	43.2 &	44.0 &	38.9 &	35.6 &	42.3 &	44.6 &	39.7 &	39.7 &  40.2 &	32.8 &	35.5 &	39.2 \\
Pavllo \etal \cite{pavllo2019} ($f=243$) & CVPR'19                & -    & -     & -    & -     & -     & -     & -    & -      & -    & -     & -     & -    & -      & -    & -      & 37.2    \\
Liu \etal \cite{Liu_2020_CVPR} ($f=243$) & CVPR'20                  & \color{blue} 34.5 & 37.1  & 33.6 & 34.2  & 32.9  & 37.1  & 39.6 & 35.8   & 40.7 & 41.4  & 33.0  & 33.8 &  33.0  & 26.6 & 26.9   & 34.7    \\
SRNet \cite{zeng2020srnet_ECCV} &  ECCV'20                        & 34.8 & \color{red} 32.1  & \color{red}28.5 & \color{red}30.7  & \color{blue}31.4  & \color{blue}36.9  & \color{blue}35.6 & \color{red}30.5   & \color{blue}38.9 & \color{blue}40.5  & \color{blue}32.5  & \color{red}31.0 & \color{blue} 29.9   & \color{red}22.5 & \color{blue}24.5   & \color{blue}32.0    \\ 
Chen \etal \cite{chen2020anatomy} ($f=243$) & TCSVT'21               & -    & -     & -    & -     & -     & -     & -    & -      & -    & -     & -     & -    & -      & -    & -      & 32.3    \\ \hline
\multicolumn{1}{l|}{\textbf{PoseFormer} ($f=81$)}  &  & \color{red}30.0 & \color{blue}33.6  & \color{blue}29.9 & \color{blue}31.0  & \color{red}30.2  & \color{red}33.3  & \color{red}34.8 & \color{blue}31.4   & \color{red}37.8 & \color{red}38.6  & \color{red}31.7  & \color{blue}31.5 & \color{red}29.0   & \color{blue}23.3 & \color{red}23.1   & \color{red}31.3    \\ 

\end{tabular}
}
\label{tab: h36mgt}
\end{table*}

\begin{table}[]
\scriptsize
\centering
\caption{Quantitative comparison with previous methods on MPI-INF-3DHP. The best scores are marked in bold.}
{
\begin{tabular}{l|c|ccc}

                                  & {} & PCK $\uparrow$ & AUC $\uparrow$ & MPJPE $\downarrow$  \\ \hline
Mehta \etal \cite{MPIINF} & 3DV'17                 & 75.7 & 39.3 & 117.6 \\
Mehta \etal \cite{mehta2017vnect} & ACM ToG'17             & 76.6 & 40.4 & 124.7 \\
Pavllo \etal \cite{pavllo2019} (81 frames)   & CVPR'19  & 86.0 & 51.9 & 84.0  \\
Pavllo \etal \cite{pavllo2019} (243 frames)   & CVPR'19 & 85.5 & 51.5 & 84.8  \\
Lin \etal \cite{lin2019trajectory} (25 frames) & BMVC'19                  & 83.6 & 51.4 & 79.8  \\ 
Li \etal \cite{Li_2020_CVPR} & CVPR'20   
               & 81.2 & 46.1 & 99.7  \\
Chen \etal \cite{chen2020anatomy} (81 frames)  & TCSVT'21   & 87.9 & 54.0 & 78.8  \\ \hline
\textbf{PoseFormer} (9 frames)                            &     &  \textbf{88.6}  & \textbf{56.4} & \textbf{77.1}  \\ 
\end{tabular}
}
\label{tab:mpi}
\end{table}

\begin{figure*}[htp]
\vspace{-5pt}
  \centering
  \includegraphics[width=0.85\linewidth]{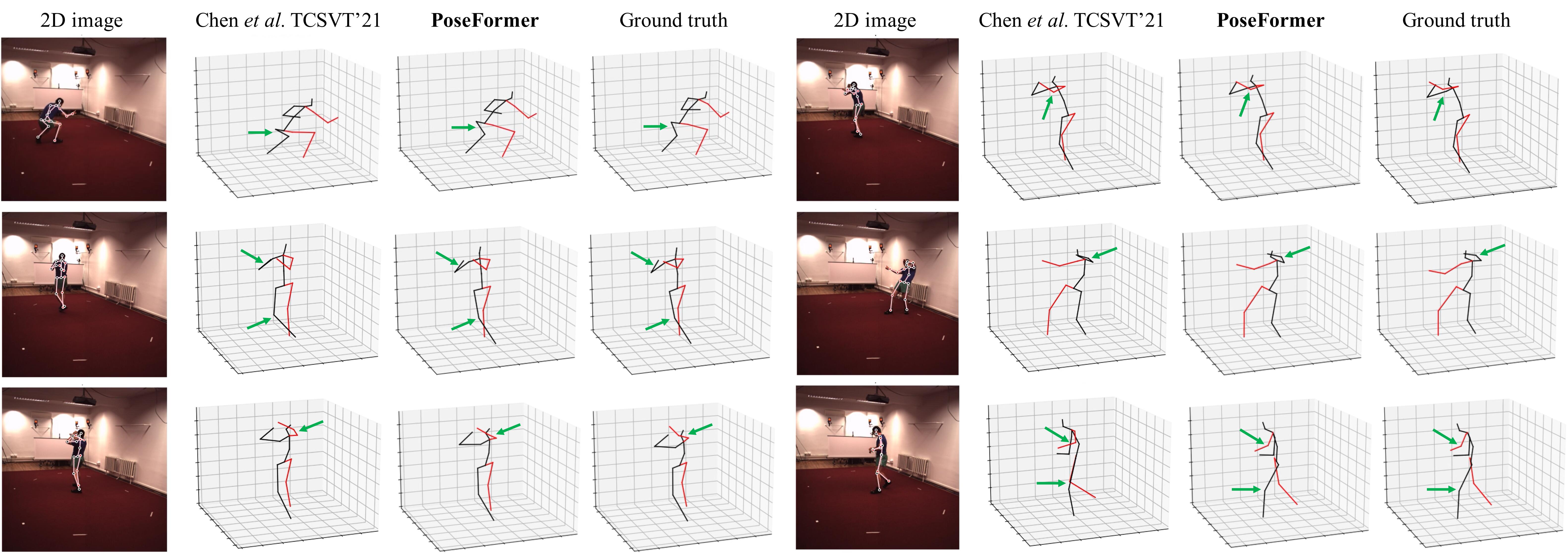}
  \caption{Qualitative comparison between our method (PoseFormer) and the SOTA approach Chen \etal \cite{chen2020anatomy} on Human3.6M test set S11 with the \textit{Photo} action. The green arrows highlight locations where PoseFormer clearly has better results.}
  \label{fig:posevisual}
  \vspace{-5pt}
\end{figure*}

\textbf{Qualitative Results.} We also provide a visual comparison between the 3D estimated pose and the ground truth.
We evaluate PoseFormer on the Human3.6M test set S11 with the \textit{Photo} action, which is one of the most challenging actions (all methods have a high MPJPE). Compared with state-of-the-art method \cite{chen2020anatomy}, our PoseFormer achieves more accurate predictions as shown in Fig. \ref{fig:posevisual}.

\subsection{Ablation Study} \label{Ablation}

To verify the contribution of the individual components of PoseFormer and the impact of hyperparameters on performance, we conduct extensive ablation experiments with the Human3.6M dataset under protocol 1. 

\begin{table}[]
\scriptsize
\centering
\caption{Ablation study on different components in PoseFormer. The evaluation is performed on Human3.6M (Protocol 1) using detected 2D pose as input. (T: Temporal only; S-T: Spatial-temporal)}
  {\begin{tabular}{c|c|c|c|c|c} 
\hline
\begin{tabular}[c]{@{}c@{}} Input length ($f$)~\end{tabular} & \begin{tabular}[c]{@{}c@{}}T\end{tabular} & \begin{tabular}[c]{@{}c@{}}S-T\end{tabular} & \begin{tabular}[c]{@{}c@{}}Spatial \\Pos Emb~\end{tabular} & \begin{tabular}[c]{@{}c@{}}Temporal\\Pos Emb\end{tabular} & MPJPE  \\ 
\hline
9                                                          & $\cmark$                                                          & $\xmark$                                                                     & $\xmark$                                                        & $\cmark$                                                      & 52.5   \\ 

9                                                          & $\xmark$                                                          & $\cmark$                                                                    & $\xmark$                                                         & $\xmark$                                                        & 51.6   \\ 

9                                                          & $\xmark$                                                           & $\cmark$                                                                   & $\cmark$                                                       & $\xmark$                                                       & 50.7   \\ 

9                                                          & $\xmark$                                                           & $\cmark$                                                                    & $\xmark$                                                         & $\cmark$                                                       & 50.5   \\ 

9                                                          & $\xmark$                                                           & $\cmark$                                                                   & $\cmark$                                                      & $\cmark$                                                       & \textbf{49.9}   \\
\hline
\end{tabular}}
\label{tab: ab1}
\vspace{-5pt}
\end{table}

\textbf{The Design of PoseFormer}. We investigate the impact of the spatial transformer, as well as the positional embeddings of the spatial and temporal transformers in Table \ref{tab: ab1}. We input 9 frames of CPN-detected 2D poses ($J=17$) to predict the 3D pose. All the architecture parameters are fixed for fairly comparing the impact of each module; the spatial transformer embedding dimension is $17 \times 32 = 544$ and the number of spatial transformer encoder layers is 4. For the temporal transformer, the embedding dimension is consistent with the spatial transformer (that is 544) and we also apply 4 temporal transformer layers. To verify the impact of our spatial-temporal design, we first compare with the transformer baseline we started with in Sec. \ref{baseline}. The results in Table \ref{tab: ab1} demonstrate our spatial-temporal transformer makes a significant impact (from 52.5 to 49.9 MPJPE), as the joint-wise correlations are more strongly modeled. This is also consistent with the results (Baseline \textit{vs.} PoseFormer) in Table \ref{tab: h36mp1} when $f=81$. Next, we evaluate the impact of the positional embeddings. We explore the four possible combinations: without positional embeddings, spatial positional embedding only, temporal positional embedding only, and both spatial and temporal positional embeddings. Comparing the results of these combinations, it is obvious that positional embeddings improve the performance. By applying these on both the spatial and temporal modules, the best overall result is achieved.

\begin{table}[]
\footnotesize
\centering
\caption{Ablation study on different architecture parameters in PoseFormer. The evaluation is performed on Human3.6M (Protocol 1) using detected 2D pose as input. $c$ is the spatial transformer patch embedding dimension. $L_S$ and $L_T$ indicate the number of layers in the spatial and temporal transformers, respectively. }
\resizebox{\linewidth}{!}
{
\begin{tabular}{c|ccccccccc}
\hline
$c$           & 16 & 16 & 16 & 32 & 32   & 32 & 48 & 48 & 48 \\ \hline
$L_{S}$  & 2  & 4  & 6  & 2  & 4    & 6  & 2  & 4  & 6  \\ \hline
$L_{T}$ & 2  & 4  & 6  & 2  & 4    & 6  & 2  & 4  & 6  \\ \hline
MPJPE       &  52.8  & 51.7   &  50.4  &  52.4  & \textbf{49.9} &  50.3  &  51.7  &  50.4  &  50.5  \\ \hline
\end{tabular}
}
\label{tab: ab2}
\vspace{-10pt}
\end{table}

\textbf{Architecture Parameter Analysis}. We explore the various parameter combinations to find the optimal network architecture in Table \ref{tab: ab2}. $c$ represents the embedded feature dimension in the spatial transformer and $L$ indicates how many layers are used in the transformer encoder. In PoseFormer, the output of the spatial transformer is flattened and added with the temporal positional embedding to form the input of the temporal transformer encoder. Thus the embedding feature dimension in the temporal transformer encoder is $c\times J$. The optimal parameters for our model are $c = 32$, $L_{S} = 4$, and $L_{T} = 4$. 

\begin{table}[]
\footnotesize
\centering
\caption{Comparison on computational complexity, MPJPE, and inference speed (frame per second (FPS)). The evaluation is performed on Human3.6M under Protocol 1 using detected 2D pose as input. FPS is based on a single GeForce GTX 2080 Ti GPU.}
\resizebox{\linewidth}{!}{
\begin{tabular}{l|c|c|c|c|c}
\hline
                   & $f$ & Parameters (M) & FLOPs (M)  & MPJPE & FPS  \\ \hline
Hossain and Little \cite{Hossain_2018_ECCV} & -     & 16.95     & 33.88 & 58.3  & -    \\
Pavllo \etal  \cite{pavllo2019}    & 27    & 8.56      & 17.09 & 48.8  & 1492 \\
Pavllo \etal   \cite{pavllo2019}   & 81    & 12.79     & 25.48 & 47.7  & 1121 \\
Pavllo \etal \cite{pavllo2019}     & 243   & 16.95     & 33.87 & 46.8  & 863  \\
Chen \etal    \cite{chen2020anatomy}    & 27    & 31.88     & 61.7  & 45.3  & 410  \\
Chen \etal    \cite{chen2020anatomy}    & 81    & 45.53     & 88.9  & 44.6  & 315  \\
Chen \etal   \cite{chen2020anatomy}     & 243   & 59.18     & 116   & 44.1  & 264  \\ \hline
\textbf{PoseFormer}               & 9     & 9.58      & 150   & 49.9      &   320   \\
\textbf{PoseFormer}               & 27    & 9.59      & 452    & 47.0      &  297    \\
\textbf{PoseFormer}               & 81    & 9.60      & 1358   &  44.3     &  269    \\ \hline
\end{tabular}
}
\label{tab: ab_complexity}
\vspace{-15pt}
\end{table}

\textbf{Computational Complexity Analysis}.
We report the model performance, total number of parameters, and estimated floating-point operations (FLOPs) per frame, and the number of output frames-per-second (FPS) with various input sequence lengths ($f$) in Table \ref{tab: ab_complexity}. Our model achieves better accuracy when the sequence length is increased, and the total number of parameters does not increase much. This is because the number of frames only affects the temporal positional embedding layer, which does not require many parameters. Compared with other models, our model requires fewer total parameters with competitive performance. We report the inference FPS of different models on a single GeForce RTX 2080 Ti GPU, following the same settings in \cite{chen2020anatomy}. Although our model's inference speed is not the absolute fastest, the speed is still acceptable for real-time inference. For complete 3D HPE processing, the 2D pose is first detected by the 2D pose detector, then the 3D pose is estimated by our method. The FPS for the common 2D pose detector is usually below 80, which means the inference speed of our model will not be the bottleneck.

\begin{figure}[htp]
  \centering
  \includegraphics[width=0.8\linewidth]{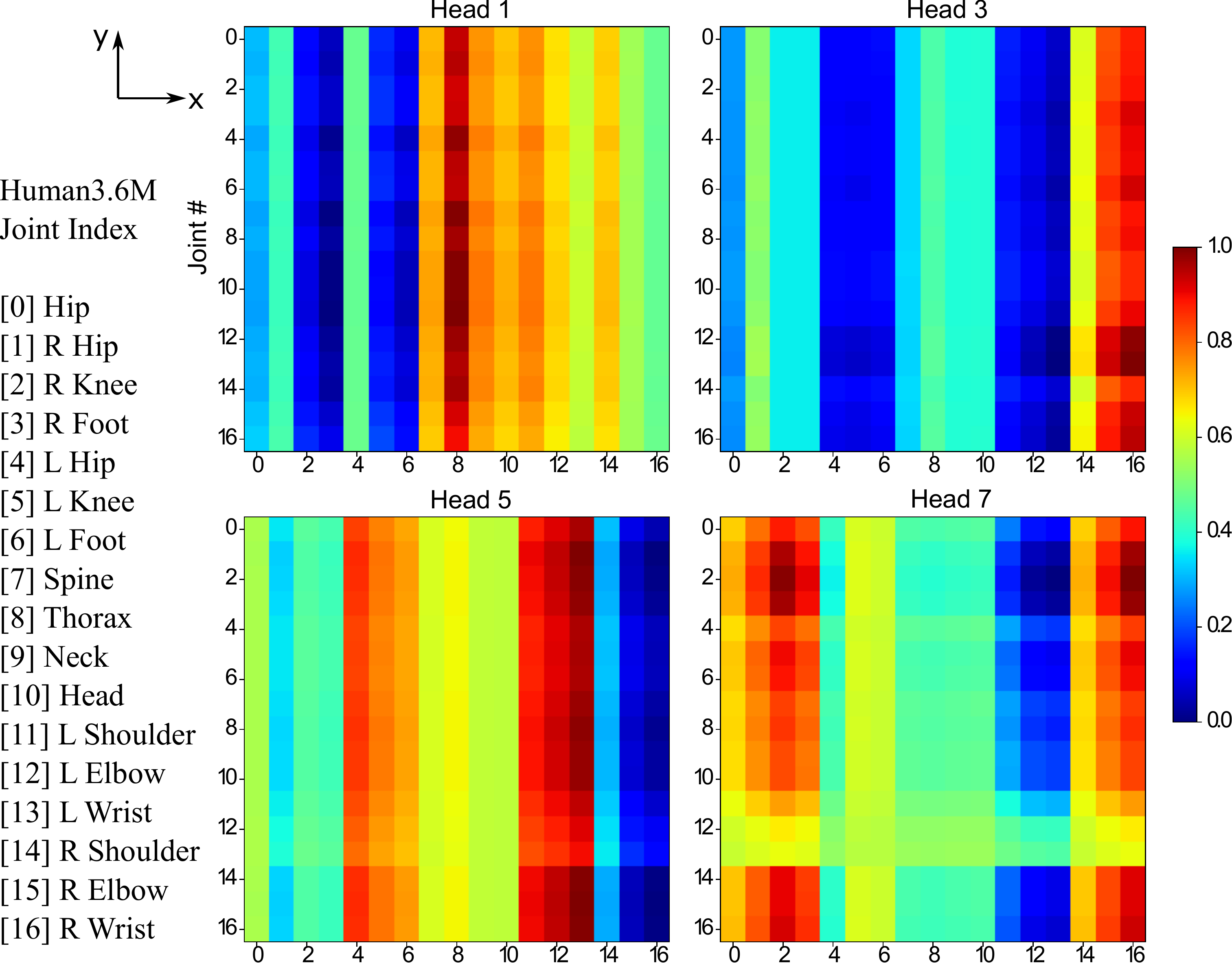}
  \vspace{-0.1em}
  \caption{Visualization of self-attentions in the spatial transformer. The x-axis (horizontal) and y-axis (vertical) correspond to the queries and the predicted outputs, respectively. The pixel $w_{i,j}$ ($i$: row, $j$: column) denotes the attention weight of the $j$-{th} query for the $i$-{th} output. Red indicates stronger attention. The attention output is normalized from 0 to 1.}
  \label{fig:s_atten}
  \vspace{-5pt}
\end{figure}

\begin{figure}[htp]
  \centering
  \includegraphics[width=0.8\linewidth]{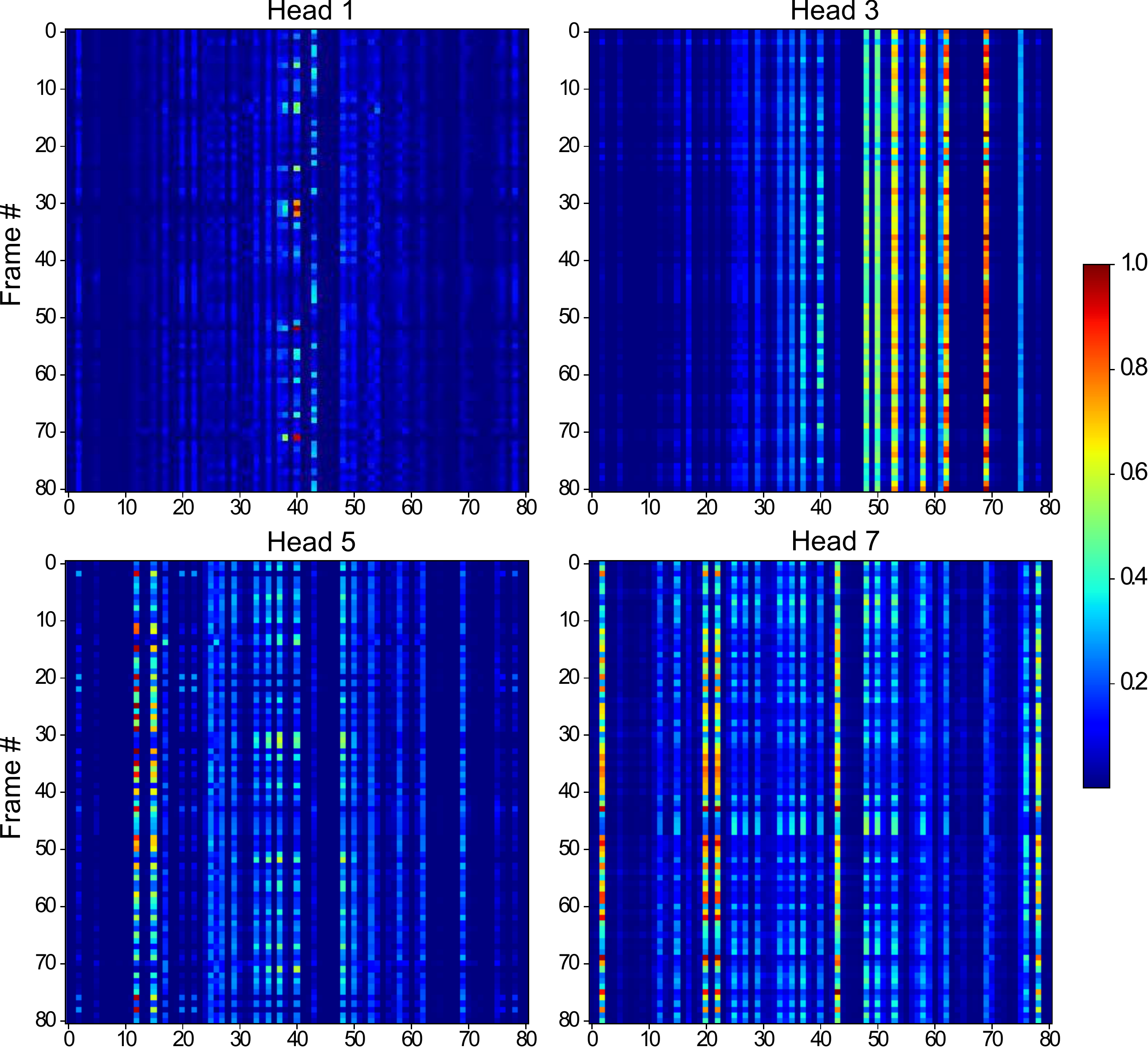}
  \vspace{-0.1em}
  \caption{Visualization of self-attentions in temporal transformer. The x-axis (horizontal) and y-axis (vertical) correspond to the queries and the predicted outputs, respectively. The pixel $w_{i,j}$ ($i$: row, $j$: column) denotes the attention weight of the $j$-{th} query for the $i$-{th} output. Red indicates stronger attention. The attention output is normalized from 0 to 1.}
  \label{fig:t_atten}
  \vspace{-5pt}
\end{figure}

\textbf{Attention Visualization}.
In order to illustrate the attention mechanism through multi-head self attention blocks, we evaluate our model on Human3.6M test set S11 for a particular action (\textit{SittingDown}) and visualize the self-attention maps from the spatial and temporal transformers separately as shown in Fig. \ref{fig:s_atten} and Fig. \ref{fig:t_atten}. For the spatial self-attention maps, the x-axis corresponds to the query of 17 joints and the y-axis indicates the attention output. As shown in Fig. \ref{fig:s_atten}, the attention heads return different attention intensities which represent the various local relations learned among the input joints. We discover that Head 3 focuses on joints 15 and 16, which are the right elbow and right wrist. Head 5 builds the connection of the left leg to the left arm (joints 4, 5, 6 and joints 11, 12, 13). These joints can be grouped as the left portion of the body, while Head 7 concentrates on the right side (joint 1, 2, 3 with joint 12, 13, 14). 

For the temporal self-attention maps in Fig. \ref{fig:t_atten}, the x-axis corresponds to the query of 81 frames and the y-axis indicates the attention output. Long term global dependencies are learned by different attention heads. The attention of Head 3 highly correlates to some frames (\eg frame 58, 62, and 69) on the right side of the center frame. Head 7 captures the dependencies of frame 1, 20, 22, 42, 78 despite their long distances. The spatial and temporal attention maps demonstrate that PoseFormer successfully models local relationships between joints, as well as captures long term global dependencies of the entire input sequence.

\begin{table}[!h]
\scriptsize
\vspace{-5pt}
\centering
\caption{MPJPE evaluation on HumanEva test set. FT indicates using pre-trained model on Human3.6M for fine tuning. }
{
\begin{tabular}{l|ccc|ccc}
\hline
              & \multicolumn{3}{c|}{walk} & \multicolumn{3}{c}{jog} \\ \hline
              & S1      & S2     & S3     & S1     & S2     & S3    \\ \hline
PoseFormer ($f=43$)    & 16.3    & 11.0     & 47.1   & 25.0   & 15.2   & 15.1  \\ \hline
PoseFormer ($f=43$) FT & \textbf{14.4}    & \textbf{10.2}   & \textbf{46.6}   & \textbf{22.7}   & \textbf{13.4}   & \textbf{13.4}  \\ \hline
\end{tabular}
}
\label{tab: ab_small}
\vspace{-5pt}
\end{table}

\textbf{Generalization to Small Datasets}. Prior work such as \cite{Dosovitskiy2020ViT} concluded that transformers do not generalize well when trained on insufficient amounts of data. We conduct an experiment with our model to investigate the transformer learning ability on a small dataset -- HumanEva \cite{HumanEva}. It is a much smaller dataset ($<$50K frames) compared with Human3.6M ($>$1M frames). Table \ref{tab: ab_small} shows the results of training from scratch as well as fine tuning using the pre-trained model on Human3.6M. We find that the performance can be improved by a large margin when fine tuning, which follows previous observations \cite{Dosovitskiy2020ViT,touvron2020deit} that transformers can perform well when pre-trained on a large-scale dataset. 

%

\section{Conclusion}

In this paper, we present PoseFormer, a pure transformer-based approach for 3D pose estimation from 2D videos. The spatial transformer module encodes the local relationships between the 2D joints and the temporal transformer module captures global dependencies across the arbitrary frames regardless of the distance. Extensive experiments show that our model achieves state-of-the-art performance on two popular 3D pose datasets.

\textbf{Acknowledgement}:  This work is partially supported by the National Science Foundation under Grant No. 1910844.

{\small
\bibliographystyle{ieee_fullname}
\bibliography{egbib}
}

\clearpage

\appendix
\section*{Appendix}
In this Appendix, we provide the following items:

\begin{itemize}

\item Comprehensive visualizations of spatial and temporal attention maps.

\item Frame-wise comparison to track the average MPJPE of all the joints across frames.

\item More qualitative comparison of estimated 3D poses. 

\item Estimated 3D poses using the proposed PoseFormer on the in-the-wild videos collected from YouTube.

\end{itemize}

We also provide demo videos to showcase the 3D human pose estimation results of our proposed PoseFormer. For more details, please visit \url{https://github.com/zczcwh/PoseFormer}

\section{Attention Visualization}
\begin{figure*}[htp]
  \centering
  \includegraphics[width=0.95\linewidth]{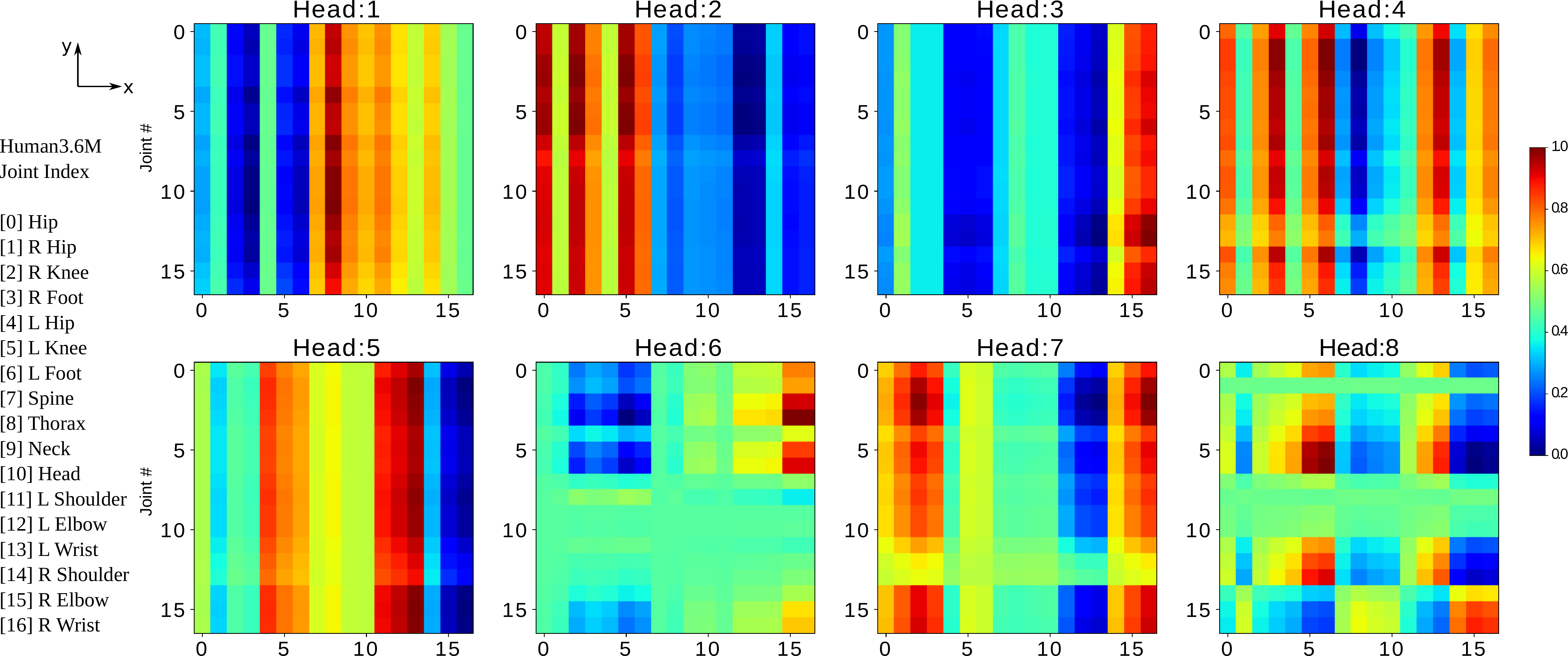}
  \caption{Visualization of self-attentions in the spatial transformer. The x-axis (horizontal) and y-axis (vertical) correspond to the queries and the predicted outputs, respectively. The pixel $w_{i,j}$ ($i$: row, $j$: column) denotes the attention weight of the $j$-{th} query for the $i$-{th} output. Red indicates stronger attention. The attention output is normalized from 0 to 1.}
  \label{fig:sup_s_atten}
\end{figure*}

\begin{figure*}[htp]
  \centering
  \includegraphics[width=1\linewidth]{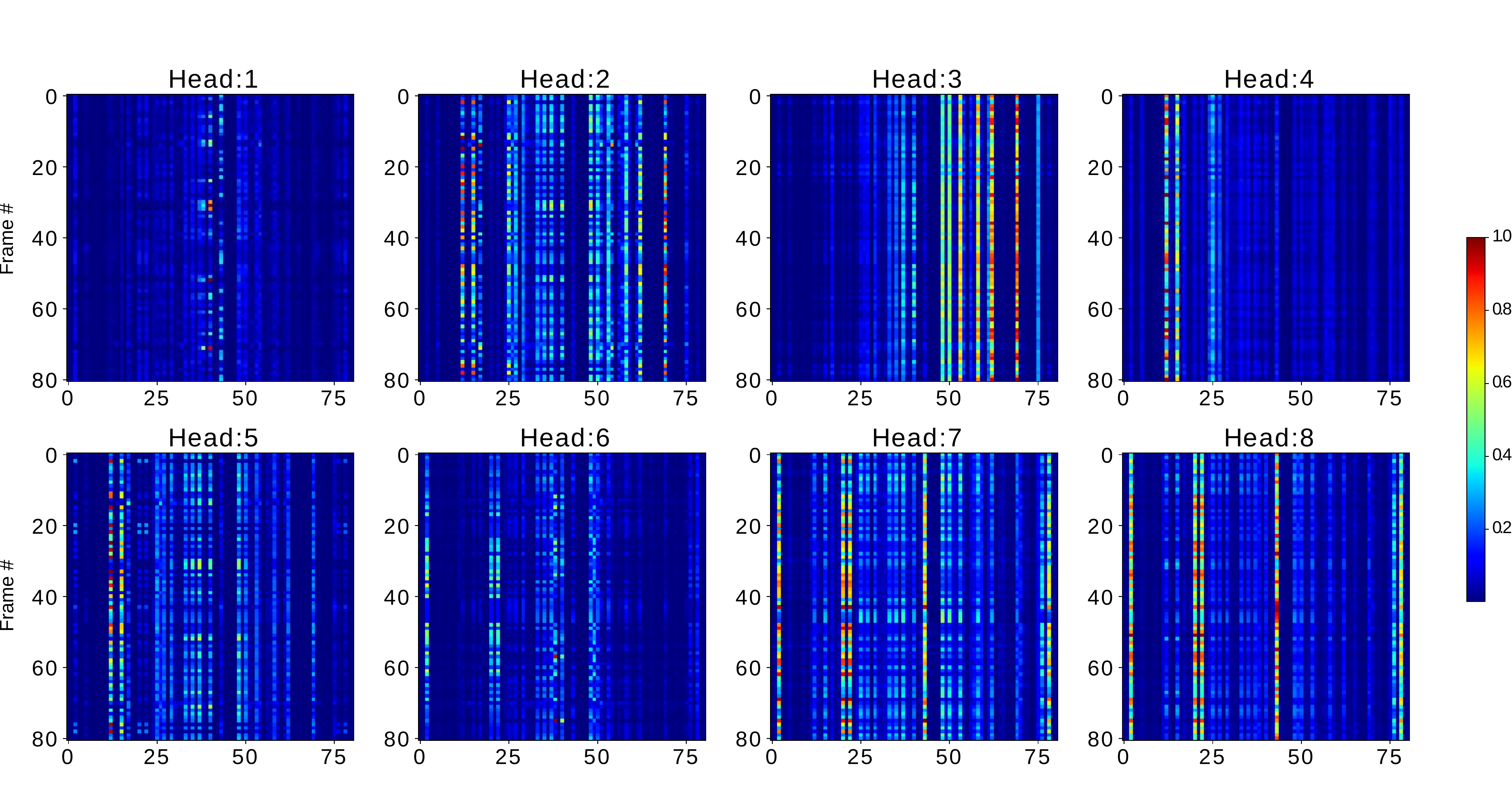}
  \caption{Visualization of self-attentions in temporal transformer. The x-axis (horizontal) and y-axis (vertical) correspond to the queries and the predicted outputs, respectively. The pixel $w_{i,j}$ ($i$: row, $j$: column) denotes the attention weight of the $j$-{th} query for the $i$-{th} output. Red indicates stronger attention. The attention output is normalized from 0 to 1.}
  \label{fig:sup_t_atten}
\end{figure*}

We present more visualization examples of spatial attention maps and temporal attention maps for all 8 heads when evaluating our PoseFormer model on Human3.6M test set S11 with the \textit{SittingDown} action. For the spatial self-attention maps in Fig. \ref{fig:sup_s_atten}, the x-axis corresponds to the query of 17 joints and the y-axis indicates the attention output. The attention heads return different attention intensities which represent the various local relations learned among the input joints. For the temporal self-attention maps in Fig. \ref{fig:sup_t_atten}, the x-axis corresponds to the query of 81 frames and the y-axis indicates the attention output. Long term global dependencies are captured by different attention heads. The spatial and temporal attention maps have demonstrated that PoseFormer successfully encodes the local relationship between 2D joints as well as models global dependencies cross the arbitrary frames regardless of the distance.

\section{Frame-wise Analysis}
\begin{figure*}[htp]
\vspace{-5pt}
  \centering
  \includegraphics[width=1\linewidth]{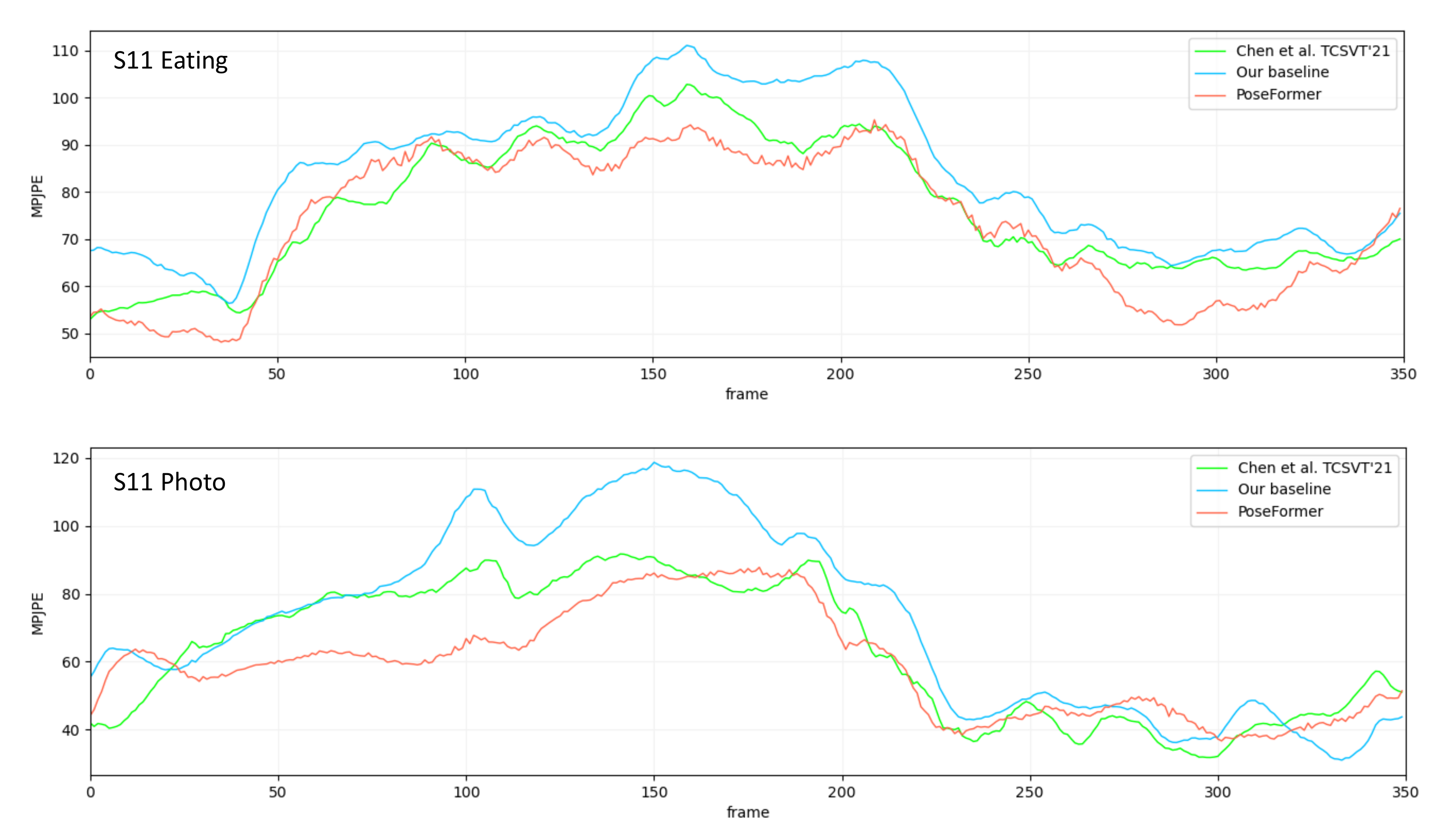}
  \caption{Frame-wise comparison between our method (PoseFormer), our baseline, and the SOTA approach Chen \etal \cite{chen2020anatomy} on Human3.6M test set. Top-figure: S11 with the \textit{Eating} action. Bottom-figure: S11 with the \textit{Photo} action.}
  \label{fig:frameerror}
  \vspace{-5pt}
\end{figure*}

We perform frame-wise estimation analysis by computing the average MPJPE of all estimated joints in each frame. As shown in Fig. \ref{fig:frameerror}, we measure the frame-wise MPJPE through Human3.6M \cite{Human3.6M} test set S11 with \textit{Eating} and \textit{Photo} actions. Our PoseFormer (red line) yields lower MPJPE in most frames of both actions, compared with our baseline (temporal transformer only) and the state-of-the-art method~\cite{chen2020anatomy}.

\section{More Qualitative Results}
\begin{figure*}[htp]
\vspace{-5pt}
  \centering
  \includegraphics[width=0.95\linewidth]{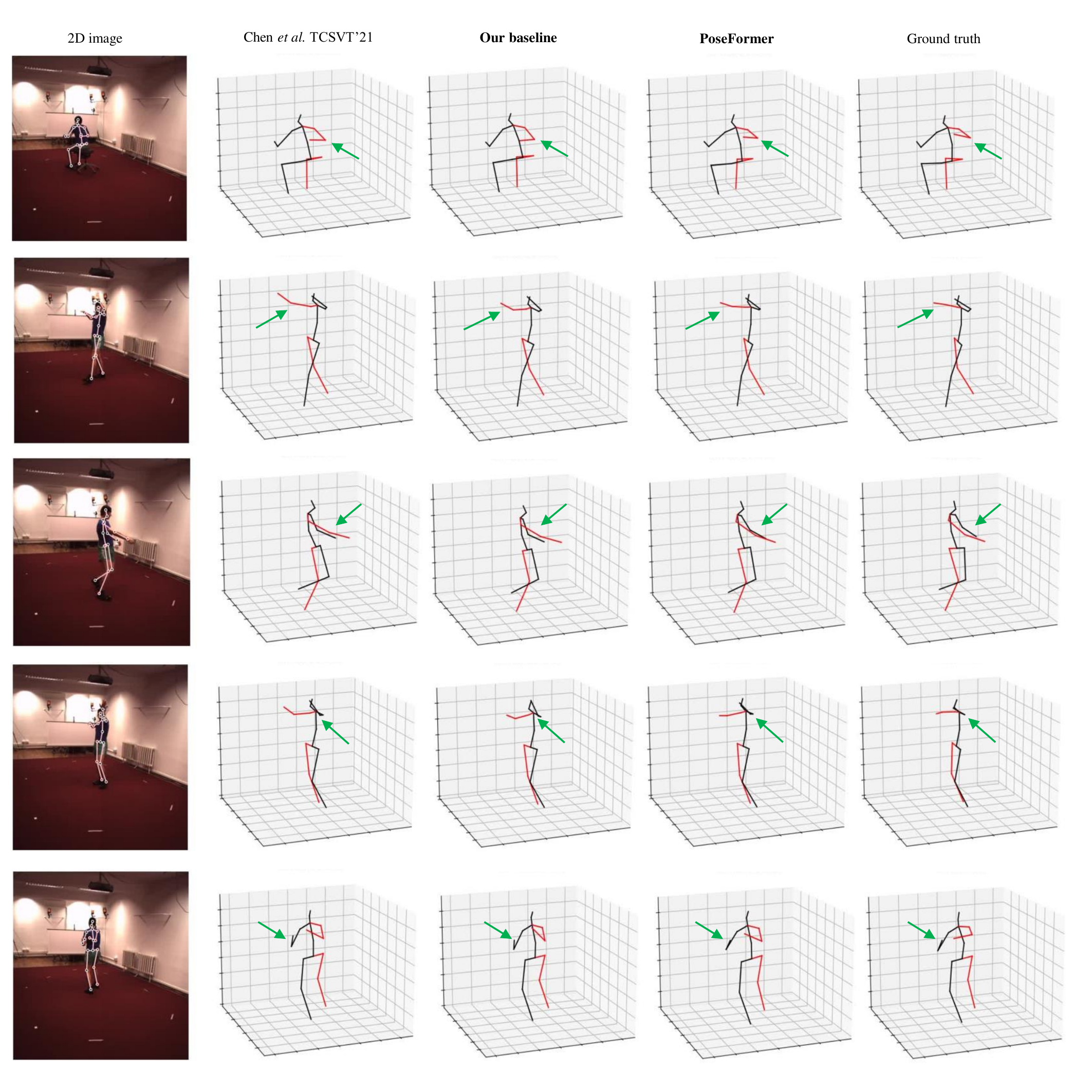}
  \caption{Qualitative comparison between our method (PoseFormer), our baseline, and the SOTA approach Chen \etal \cite{chen2020anatomy} on Human3.6M test set S11 with the \textit{Greeting} and \textit{WalkDog} actions. The green arrows highlight locations where PoseFormer clearly has better results.}
  \label{fig:sup_posevisual}
\end{figure*}

We provide more visual comparison between the 3D estimated pose and the ground truth. We evaluate PoseFormer on the Human3.6M test set S11 with the \textit{Greeting} and \textit{WalkDog} actions. Compared with the state-of-the-art method \cite{chen2020anatomy} and our baseline, PoseFormer achieves more accurate estimations as shown in Fig. \ref{fig:sup_posevisual}.

\section{Performance on Videos in-the-wild}
\begin{figure*}[htp]
\vspace{-5pt}
  \centering
  \includegraphics[width=0.95\linewidth]{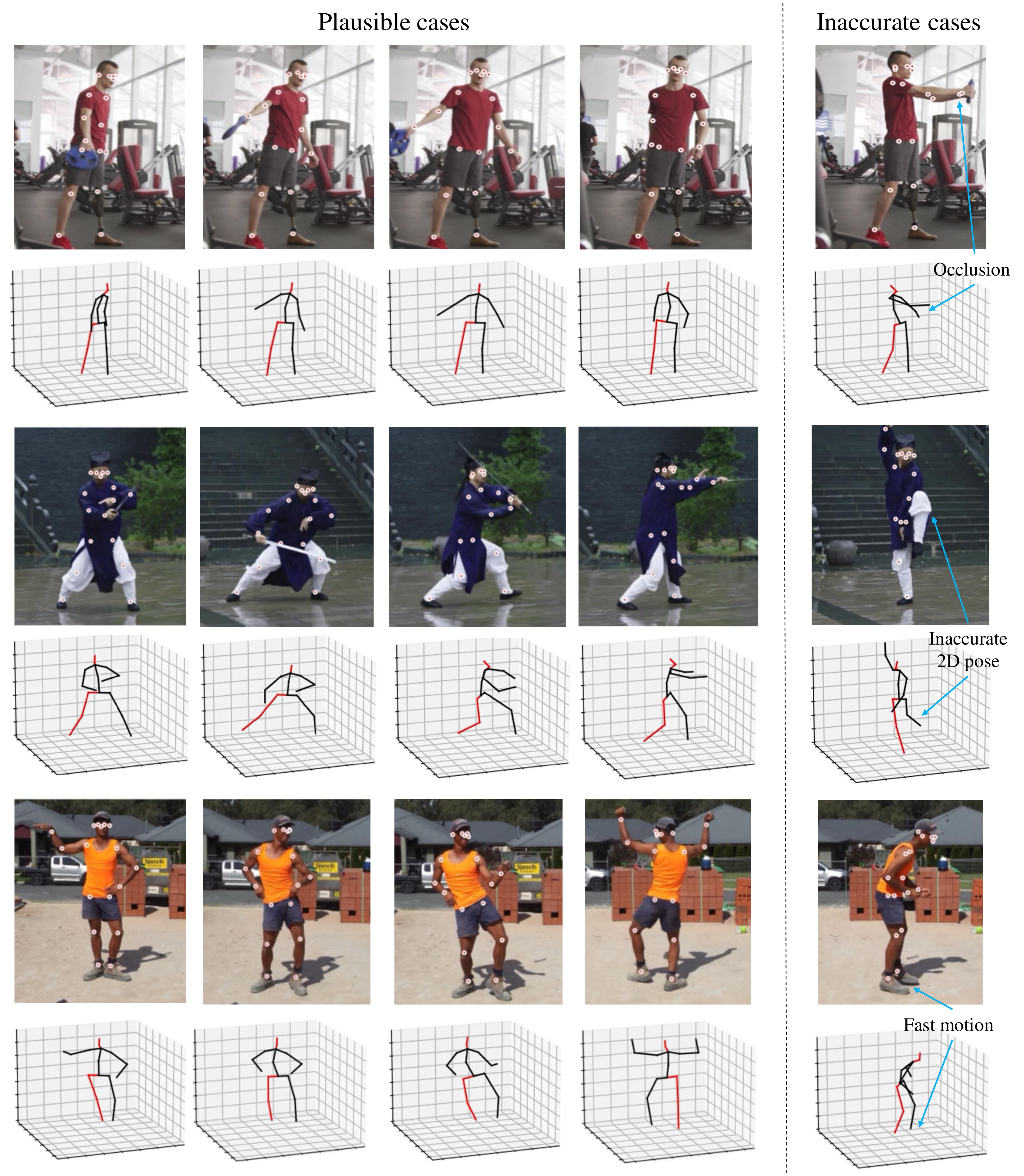}
  \caption{Qualitative results on in-the-wild videos: original frame sequence with detected 2D joints and the recovered 3D poses using PoseFormer.}
  \label{fig:wild}
  \vspace{-5pt}
\end{figure*}

Our model was trained on the indoor dataset: Human3.6M that the background is static and the camera capture setting is known. Estimating 3D human pose from in-the-wild videos is more challenging due to the dynamic environment and unknown camera setting. There are often high variations in foreground/background objects appearances and severe occlusions in unconstrained environment. We also evaluate the performance of our PoseFormer on some online videos from YouTube as shown in Fig. \ref{fig:wild}. We first use AlphaPose \cite{fang2017rmpe} as 2D pose detector to generate 2D poses from the video frames, then apply PoseFormer for 3D pose estimation. We observe that PoseFormer achieves acceptable performance in most of the frames, but there are still some failure cases (see Fig. \ref{fig:wild}) due to inaccurate 2D pose detection, occlusion, and fast motion. Since PoseFormer is a 2D-to-3D lifting approach, any incorrect detected 2D poses may lead to inaccurate 3D pose estimation. Occlusion is a key challenge remains in 3D HPE since the information is missing. Moreover, estimation from the extreme fast motion may be affected by the motion blurring of frames.

\end{document}